\documentclass[sigconf,10pt]{acmart}
\renewcommand\footnotetextcopyrightpermission[1]{} 
\usepackage{amsmath,epsfig}
\usepackage{threeparttable}
\usepackage{multirow}
\usepackage{booktabs}
\usepackage{amssymb}
\usepackage{verbatim}
\usepackage{subfigure}
\usepackage{graphicx}
\usepackage{float}
\usepackage{stfloats}
\usepackage{flushend}

\def\BibTeX{{\rm B\kern-.05em{\sc i\kern-.025em b}\kern-.08emT\kern-.1667em\lower.7ex\hbox{E}\kern-.125emX}}

\begin{document}

%
\title{MAANet: Multi-view Aware Attention Networks \\
for Image Super-Resolution}



\author{Jingcai Guo}
\affiliation{%
 \institution{Hong Kong Polytechnic University}
 \state{Hong Kong SAR}
 \country{China}}
\email{cscjguo@comp.polyu.edu.hk}

\author{Shiheng Ma}
\affiliation{%
 \institution{Shanghai Jiao Tong University}
 \state{Shanghai}
 \country{China}}
\email{ma-shh@sjtu.edu.cn}

\author{Song Guo}
\affiliation{%
 \institution{Hong Kong Polytechnic University}
 \state{Hong Kong SAR}
 \country{China}}
\email{song.guo@polyu.edu.hk}

\begin{abstract}
In most recent years, deep convolutional neural networks (DCNNs) based image super-resolution (SR) has gained increasing attention in multimedia and computer vision communities, focusing on restoring the high-resolution (HR) image from a low-resolution (LR) image. However, one nonnegligible flaw of DCNNs based methods is that most of them are not able to restore high-resolution images containing sufficient high-frequency information from low-resolution images with low-frequency information redundancy. Worse still, as the depth of DCNNs increases, the training easily encounters the problem of vanishing gradients, which makes the training more difficult. These problems hinder the effectiveness of DCNNs in image SR task. To solve these problems, we propose the Multi-view Aware Attention Networks (MAANet) for image SR task. Specifically, we propose the local aware (LA) and global aware (GA) attention to deal with LR features in unequal manners, which can highlight the high-frequency components and discriminate each feature from LR images in the local and the global views, respectively. Furthermore, we propose the local attentive residual-dense (LARD) block, which combines the LA attention with multiple residual and dense connections, to fit a deeper yet easy to train architecture. The experimental results show that our proposed approach can achieve remarkable performance compared with other state-of-the-art methods.
\end{abstract}


%
%
%
\maketitle

\section{Introduction}
Image super-resolution (SR) is a hot topic that keeps rising in multimedia and computer vision \cite{glasner2009super,dong2016image,lim2017enhanced,haris2018deep}. It aims at restoring the high-resolution (HR) image from a low-resolution (LR) image via SR algorithms. Image SR is widely used in various applications, including but not limited to medical imaging \cite{oktay2018anatomically}, satellite image analysis \cite{benecki2018evaluating}, security and surveillance \cite{uiboupin2016facial,yin2018deep}, high-definition video processing \cite{huang2018video}, etc. Since LR images have lost much information compared to their HR counterparts, the image SR is an ill-posed problem that has multiple solutions for LR inputs \cite{baker2002limits,lin2004fundamental}. Recently, with the rapid development of deep learning techniques, deep learning based SR algorithms achieve superior performance over traditional SR algorithms in terms of not only quantitative metrics including peak signal-to-noise ratio (PSNR) and structural similarity index (SSIM), but also qualitative measures with more visually pleasing HR images.

The SR algorithms with deep learning can be grouped into two main-streams. One is based on generative adversarial networks (GANs), involving two neural networks that contest with each other in a zero-sum game framework. This technique helps to generate images that look at least superficially authentic to human observers. Some representative methods include SRGAN \cite{ledig2017photo} and EnhanceNet \cite{sajjadi2017enhancenet}. The other is built upon deep convolutional neural networks (DCNNs), which adopts a deep network architecture to learn a mapping from LR images to their HR counterparts with L1 or L2 loss in pixel space. Some representative methods include SRCNN \cite{dong2016image}, FSRCNN \cite{dong2016accelerating}, VDSR \cite{kim2016accurate}, LapSRN \cite{lai2017deep}, EDSR \cite{lim2017enhanced}, RDN \cite{zhang2018residual}, DBPN \cite{haris2018deep}, etc. Generally, the GANs based SR algorithms can obtain more visually pleasing images, contributing to the qualitative performance. However, they do not perform very well in quantitative performance, i.e., PSNR and SSIM. Furthermore, the training of GANs is more difficult and costly for relying on more training data and tough convergence. 

In contrast, DCNNs based SR algorithms are much easier to train and can achieve better quantitative performance in PSNR and SSIM. Meanwhile, they are also satisfactory in qualitative performance in real-world applications. In recent years, DCNNs based image SR algorithms have achieved significant improvements and obtained successive state-of-the-art performances over traditional image SR algorithms. However, most of them do not pay enough attention to the limited high-frequency information from LR images. This motivates our idea that to highlight the high-frequency components from LR inputs and to deal with LR features unequally are also of crucial importance for restoring better quality HR images.

To solve these problems, in this paper we propose the local aware (LA) and global aware (GA) attention to adaptively handle LR features in unequal manners. On the one hand, the LA attention unequally highlights the high-frequency components within each LR feature map in the local view. On the other hand, in the global view, the GA attention focuses on unequally re-weighting each feature map after the LR inputs pass through the upscale unit in the HR feature space. Furthermore, to achieve a deeper yet easy training network architecture, we propose the local attentive residual-dense (LARD) block to combine the LA attention with multiple residual and dense connections in the LR feature space. Extensive experiments show a remarkable performance improvement of our model compared with other existing methods.

Our contributions can be summarized as follows. (1) We propose the local aware (LA) attention to highlight the high-frequency components from LR inputs. (2) We propose the global aware (GA) attention to deal with each LR features and re-weight them in unequal manners. (3) We propose the LARD block to construct a very deep and trainable network architecture.

\begin{figure*}[t]
    \centerline{\includegraphics[width=0.89\textwidth]{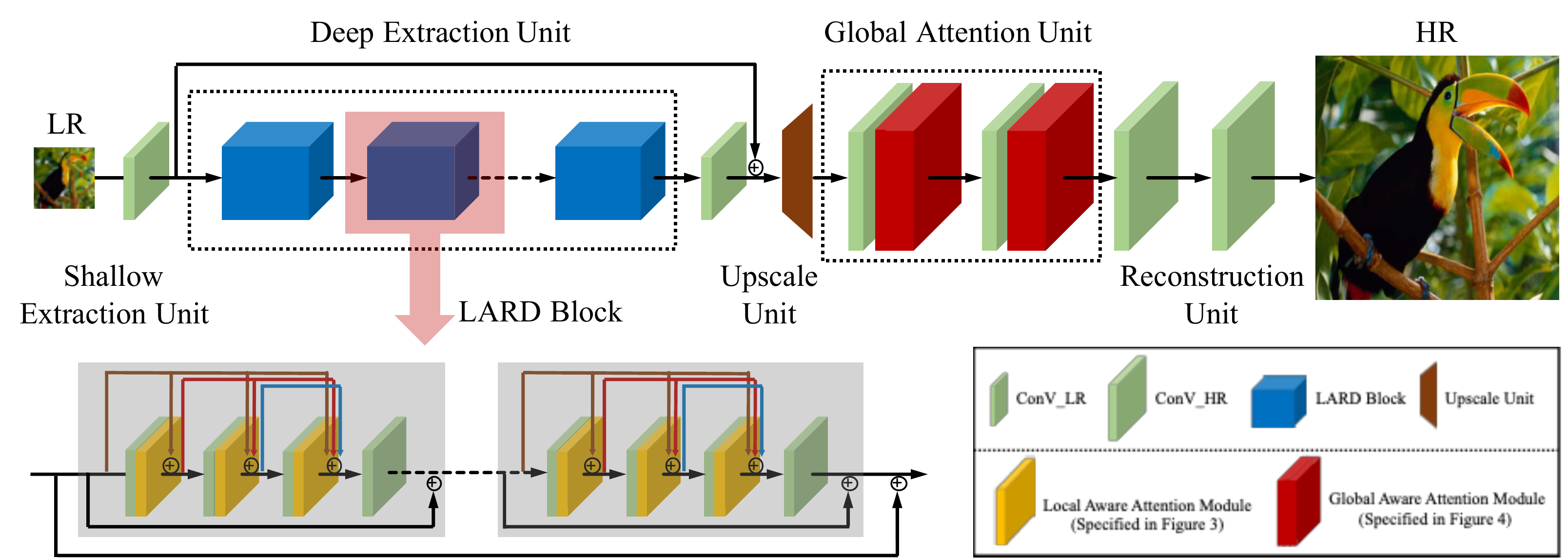}}
    \caption{Framework of MAANet}
    \label{fig2}
\end{figure*}
\section{Related Work} 

\subsection{Super-Resolution with DCNNs}
Recently, traditional super-resolution (SR) algorithms \cite{li2001new,zhang2006edge,sun2008image,tai2010super,yang2013fast} are severely inferior to multiple DCNNs based SR algorithms. Among them, SRCNN \cite{dong2016image} is the pioneering work of DCNNs for image SR task. It first uses bicubic to scale up the LR inputs to HR size and then outperforms traditional image SR algorithms by using only three convolutional layers. Extended from SRCNN, FSRCNN \cite{dong2016accelerating} with a deeper network architecture obtains faster speed and better results by scaling up LR inputs to HR size at the last few layers. VDSR \cite{kim2016accurate} increases the network depth to a very high level and demonstrates the depth of networks is of crucial importance to image SR task. LapSRN \cite{lai2017deep} is based on a cascade of CNNs, which progressively predicts the sub-band residuals in a coarse-to-fine fashion and obtains progressive reconstruction. EDSR \cite{lim2017enhanced} removes unnecessary modules in conventional residual networks to enhance the deep super-resolution model and further improves it by expanding the model size while stabilizing the training procedure. RDN \cite{zhang2018residual} proposes to fully exploit the hierarchical features from all the convolutional layers by using residual dense connections to extract abundant local features, and by fusing them to jointly and adaptively learn global hierarchical features in a holistic way. DBPN \cite{haris2018deep} constructs mutually connected up-sampling and down-sampling stages, each of which represents different types of image degradation and high-resolution components. It iteratively exploits up-sampling and down-sampling layers, and provides an error feedback mechanism for projection errors at each stage. However, most of these methods do not pay enough attention to the limited high-frequency information from LR images. From our observation, high-frequency components refer to the pixel intensities that are rapidly changing in space. Compared to the HR images, the LR images have been ruined by losing much high-frequency information, i.e., sharp contrast edges, and leaving redundant low-frequency information, i.e., smooth gradients. If the restored images contain insufficient high-frequency information, they will appear blurring and missing details, resulting in low visual comfort. In our model, we investigate and solve these problems by our proposed MAANet.

\subsection{Attention Mechanism}
The attention mechanism stems from the study of human vision. In cognitive science, because of the bottleneck of information processing, humans selectively focus on part of the whole information while ignoring others. Similarly, in multimedia and computer vision fields, the attention mechanism is introduced to bias the allocation of available processing resources towards the most informative components of an input. In recent years, a large number of deep learning models have incorporated the attention mechanisms and achieved promising results. Ba et al. \cite{ba-attention-2015} extends the attention-based RNN model to multiple objects detection task that learns to localize and recognize multiple objects despite being given only class labels. Some employ attention mechanisms in the visual question answering task, such as generating question-guided attention to image feature maps for each question \cite{chen2015abc}, the question-guided spatial attention to images for questions of spatial inference \cite{xu2016ask} and querying an image and inferring the answer multiple times to narrow down the attention to images progressively via stacked attention networks \cite{yang2016stacked}. Wang et al. \cite{wang2017residual} proposes the residual attention network that composed of multiple attention modules, and can be easily scaled up to hundreds of layers with superior performance. Hu et al. \cite{hu2018squeeze} proposes the squeeze-and-excitation block combined with attention, which can improve the representational power of a network by explicitly modeling the interdependencies between the channels of its convolutional features. Yan et al. \cite{yan2019dual} uses the attention mechanism combined with a proposed semantic-aware meta-learning framework, in which the model explicitly incorporates class sharing across tasks and focuses on only semantically informative parts of input images in each task for few-shot learning. Despite the progress made, however, the attention mechanism has rarely been utilized in some low-level vision tasks such as image super-resolution. In this paper, we propose to apply attention mechanisms to adaptively handle the input low-resolution features in unequal manners with both local and global views.

\section{Proposed Method}

\subsection{Networks Architecture}
As shown in Figure \ref{fig2}, our MAANet mainly consists of five parts: (1) Shallow extraction unit is exactly a convolution layer (ConV). (2) Deep extraction unit contains several local attentive residual-dense (LARD) blocks in series. It can extract rich deep features with sufficient high-frequency information by combining the LA attention with multiple residual and dense connections. (3) Upscale unit resizes the low-resolution (LR) input to high-resolution (HR) scale. (4) Global attention unit unequally re-weights each feature map in the HR feature space. (5) Reconstruction unit restores the HR output. Considering an input LR image $I_{LR}$, we denote its HR counterpart as $I_{SR}$. In our MAANet, we first input $I_{LR}$ to the shallow feature extraction unit and obtain its shallow feature
\begin{equation}
F_{S} = E_{S}\left ( I_{LR} \right ),
\end{equation}
where $E_{S}\left ( \cdot \right )$ and $F_{S}$ denote the shallow feature extraction unit and the shallow feature, respectively. Then $F_{S}$ is further inputted to the deep feature extraction unit as
\begin{equation}
F_{D} = E_{D}\left ( F_{S} \right ),
\end{equation}
where $F_{D}$ denotes the deep feature and $E_{D}\left ( \cdot \right )$ denotes the deep feature extraction unit with $k$ LARD blocks. This unit helps to construct a very deep and trainable network architecture. Next, the obtained deep feature $F_{D}$ continues passing through the upscale unit $E_{U}\left ( \cdot \right )$ and resizing to the HR feature maps, i.e., 
\begin{equation}
F_{U} = E_{U}\left ( F_{D} \right ).
\end{equation}
The upscaled $F_{U}$ is still on a deep and general stage \cite{yosinski2014transferable}, so we can further apply the global attention unit to discriminate and re-weight each feature map in the HR feature space. Then we can have
\begin{equation}
F_{G} = E_{G}\left ( F_{U} \right ),
\end{equation}
where $E_{G}\left ( \cdot \right )$ and $F_{G}$ are the global attention unit and the obtained feature, respectively. Last, the reconstruction unit $E_{R}\left ( \cdot \right )$ restores the HR output as 
\begin{equation}
I_{SR} = E_{R}\left ( F_{G} \right ) = \Phi _{MAANet}\left ( I_{LR} \right ),
\end{equation}
where $\Phi _{MAANet}\left ( \cdot \right )$ denotes the trained networks that map an LR input image $I_{LR}$ to its HR counterpart image $I_{SR}$. The detailed layer and parameter settings for each unit are specified in Section 4. 

Similar to most DCNNs based SR algorithms \cite{dong2016image,dong2016accelerating,kim2016accurate,lai2017deep,lim2017enhanced,zhang2018residual,haris2018deep}, we also adopt $L_{1}$ loss to optimize our MAANet. Given a set of training data $\mathcal{D}_{Tr} = \left \{ I_{LR}^{i}, I_{SR}^{i} \right \}_{i=1}^{n}$, which contains $n$ LR input images and their corresponding HR counterparts. The loss function is defined as:
\begin{equation}
\mathcal{L}\left ( \theta  \right ) = \frac{1}{n}\sum_{i=1}^{n}\left \| \Phi _{MAANet}\left ( I_{LR} \right ) - I_{SR} \right \|_{1},
\end{equation}
where $\left \| \cdot \right \|_{1}$ denotes the $L_{1}$ norm and $\theta$ denotes the parameter set of our model. 

\begin{figure*}[t]
    \centerline{\includegraphics[width=0.87\textwidth]{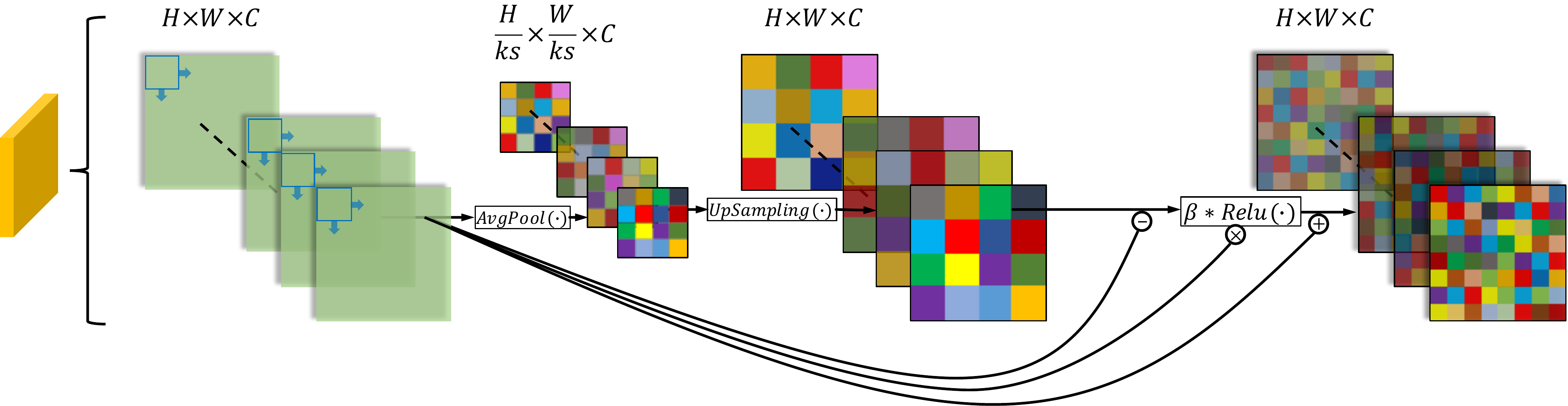}}
    \caption{Local aware attention}
    \label{fig3}
\end{figure*}

\subsection{Local Aware Attention}
The local aware (LA) attention module is a key building block in our model. Previous DCNNs based SR algorithms normally treat each feature from LR inputs equally in their network layers. However, based on our observation, the LR image has lost much high-frequency information, e.g., sharp contrast edges, textures, etc., compared with its HR counterpart. The LR inputs only contain limited high-frequency components which make the restored HR images visually uncomfortable. Worse still, with low-frequency information redundancy, the SR algorithms are prone to generate overly smooth images to cater to higher quantitative results such as PSNR and SSIM metrics. Therefore, it is of crucial importance to pay more attention to the limited high-frequency information from LR inputs.

Based on these analyses, we propose to deal with each LR feature in unequal manners. First, we deal with LR inputs in the LR feature space. The LA attention module aims to highlight the high-frequency components in the local view, i.e., within each LR feature map in the LR feature space. As shown in Figure \ref{fig3}, we now give more details about the structure of LA attention module. Consider a layer in the deep feature extraction unit, we have a tensor $T_{D}$ with size $H \times W \times C$, which denotes that there are $C$ feature maps in $C$ channels, and each feature map has the height $H$ and the width $W$. We first apply the average pooling to $T_{D}$, i.e., 
\begin{equation}
T_{DA} = AvgPool(T_{D}, ks, s),
\end{equation}
where $ks$ and $s$ are the pooling kernel size and stride, respectively. In this step we set $s = ks$, so that the obtained tensor $T_{DA}$ is with size $\frac{H}{ks} \times \frac{W}{ks} \times C$, and each value in $T_{DA}$ represents the average intensity of a specific sub-region in the corresponding feature map of $T_{D}$. Next, by using up-sampling with a scale parameter equal to $ks$, we can obtain an $H \times W \times C$ tensor $T_{DU}$ that has the same size as $T_{D}$:
\begin{equation}
T_{DU} = UpSampling(T_{DA}, ks).
\end{equation}
The obtained $T_{U}$ can be regarded as an expression of the average smoothness information of the sub-regions in the original $T_{D}$. Essentially, each element in the feature map represents the embedded feature and signal intensity of a specific region in the feature map of the previous layer. So, in order to highlight the high-frequency information in the local view, i.e., the sub-regions of each feature map, we can have an alternative strategy to highlight the elements in each feature map of the next layer. For this purpose, we subtract $T_{DU}$ from $T_{D}$ in element-wise and activate the remainings, i.e.,  
\begin{equation}
T_{DR} = Relu\left ( T_{D} - T_{DU} \right ),
\end{equation}
where $Relu\left ( \cdot \right )$ is the rectified linear unit which takes the positive part of its argument, and $T_{DR}$ is the regional average residual. Each value in $T_{DR}$ denotes whether its corresponding sub-region in the feature map of the previous layer is beyond the average sub-smoothness or not. Last, two shortcuts from $T_{D}$ to $T_{DR}$ (element-wise multiplication), and to the very end (element-wise sum) are constructed as:
\begin{equation}
\begin{aligned}
\hat{T_{D}} &= T_{D} + \beta T_{DR} \otimes T_{D},
\end{aligned}
\end{equation}
where $\beta$ is a tiny hyper-parameter which controls the degree of highlighting and $\otimes$ denotes the element-wise multiplication. The updated $\hat{T_{D}}$ is then inputted to the next layer of the deep feature extraction unit.

\subsection{Global Aware Attention}
In contrast to the LA attention, the global aware (GA) attention handles the global view of the LR inputs, i.e., unequally discriminating and re-weighting each feature map after passing through the upscale unit in the HR feature space (Figure \ref{fig2}). Conventional DCNNs based SR algorithms treat each feature map equally in their networks. However, when performing convolution operations, different filters produce different types of features. These features then form different feature maps to represent an input data in multi-level aspects. In other words, the representation ability of each feature map varies from one to another. Therefore, it is necessary to pay more attention to discriminate these feature maps in our model.

\begin{figure}[t]
    \centerline{\includegraphics[width=0.47\textwidth]{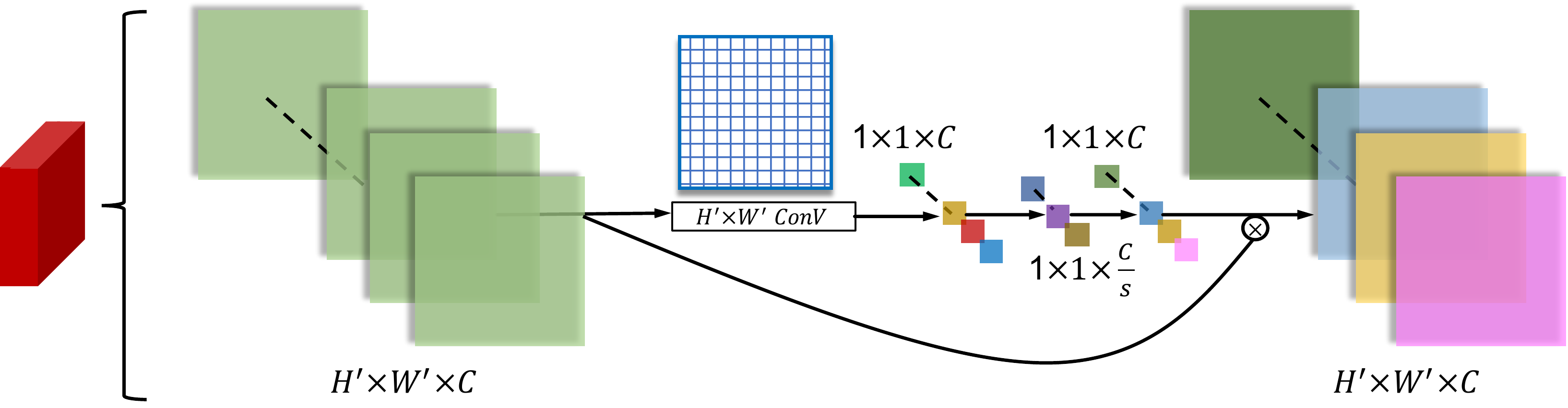}}
    \caption{Global aware attention}
    \label{fig4}
\end{figure}

Based on these analyses, we propose the global aware attention unit as shown in Figure \ref{fig4}. We consider an ${H}' \times {W}' \times C$ tensor $T_{G}$ with the same height ${H}'$ and width ${W}'$ as HR images. First, we perform a convolution operation with kernel size ${H}' \times {W}'$ to $T_{G}$, and we can have
\begin{equation}
V_{G} = ConV\left ( T_{G}, [{H}',{W}'] \right ).
\end{equation}
The ${H}' \times {W}'$ filter generates a $1 \times 1 \times C$ tensor $V_{G}$ for $T_{G}$. Each element in $V_{G}$ denotes some statistics information of each feature map of $T_{G}$. Unlike existing attention models that use global average pooling, we use a trainable ${H}' \times {W}'$ filter to obtain this statistics information, making the model more robust and expressive. Then by further using two convolution operations to the obtained $V_{G}$, we can have
\begin{equation}
{V_{G}}' = \sigma\left ( ConV_{2}\left ( Relu\left ( ConV_{1}\left (V_{G},C,{C}'  \right ) \right ),{C}',C  \right ) \right ),
\end{equation}
where $ConV_{1}$ is the first convolution operation that shrinks $V_{G}$ from $C$ to ${C}'$ channels (${C}' < C$), and the second one, i.e., $ConV_{2}$, upscales it back. $Relu\left ( \cdot \right )$ and $\sigma\left ( \cdot \right )$ are the rectified linear unit and sigmoid gateway, respectively. The updated ${V_{G}}'$ can represent a more general global statistics for each feature map in the HR feature space. Last, a shortcut from $T_{G}$ to the updated ${V_{G}}'$ is constructed, and the original $T_{G}$ is re-weighted as:
\begin{equation}
\hat{T_{G}} = {V_{G}}' \cdot T_{G}.
\end{equation}
With the GA attention, our MAANet can adaptively pay unequal attention to each feature map in the global view. 

\subsection{LARD Block}

\begin{figure}[t]
    \centerline{\includegraphics[width=0.42\textwidth]{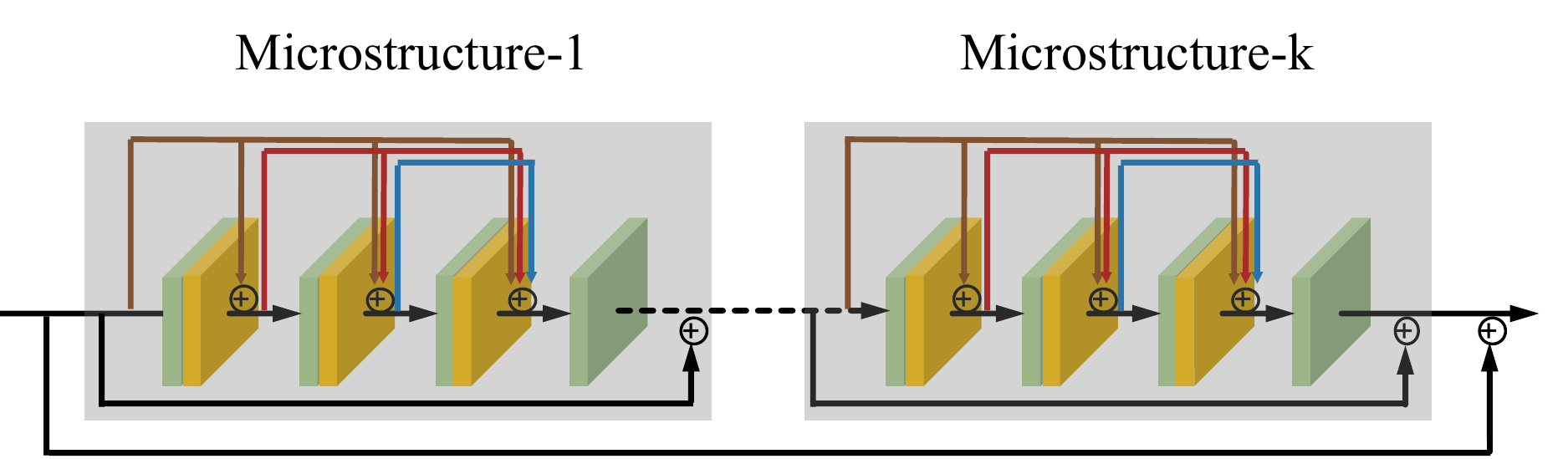}}
    \caption{LARD block}
    \label{fig5}
\end{figure}

It has been demonstrated that the depth of networks is of crucial importance to image SR task \cite{kim2016accurate}. In our MAANet, the deep extraction unit contains several local attentive residual-dense (LARD) blocks in series, as shown in Figure \ref{fig5}. This architecture allows us to easily build a deep yet easy to train networks by repeatedly stacking the LARD block.

As investigated in ResNet \cite{he2016deep}, as the depth increases, the residual learning framework can dramatically alleviate the problem of vanishing gradients and better guarantee the model's representation ability by adopting skip connections from one layer to latter layers. The DenseNet \cite{huang2017densely} further extends the skip connections to each single layer and constructs the dense connections in the network architecture. Consequently, one layer receives the feature maps of all preceding layers, resulting in two merits, i.e., deeper trainable network architecture and better cross-layer extraction ability. Inspired by these ideas, the proposed LARD block combines our local aware attention with multiple residual and dense connections, to extract rich deep features from LR inputs. The LARD block can not only highlight the high-frequency components in the local view, i.e., within the sub-regions of each feature map, but also strengthen the model's extraction and representation ability.

\section{Experiments}

\subsection{Implementation and training details}
Our MAANet mainly consists of five parts: shallow extraction unit, deep extraction unit, upscale unit, global attention unit, and reconstruction unit. We now give the implementation details of each part. 

First, the shallow extraction unit is exactly a convolution layer which contains 64 filters with the kernel size $3 \times 3$ and the stride 1. We ensure the obtained feature maps have the same height and width as the input LR images by adding zero paddings. The obtained feature maps are then inputted to the deep extraction unit, which contains multiple LARD blocks in series. It can extract rich deep features with sufficient high-frequency information. Within each LARD block, there exists three serial microstructures. As shown in Figure \ref{fig5}, each microstructure contains four convolution layers and three LA attention modules. The first three convolution layers are paired with these LA attention modules one by one. Each of these convolution layers contains 32 filters with the kernel size $3 \times 3$. The stride is set to 1 and we also use zero paddings to ensure that the obtained feature maps have the same size as LR inputs. The LA attention module has the $4 \times 4$ kernel size, and the stride is also set to 4, which makes the sub-region have the same size as the kernel. We construct dense connections among the first three pairs of convolution layer and LA attention module. The fourth convolution layer is added to the end, combined with a residual connection from the beginning to the very end of this microstructure. The residual part is scaled by $\frac{1}{d}$, where $d$ denotes the number of LA attention modules. Last, a wider residual connection from the beginning to the very end of the LARD block is constructed (Figure \ref{fig2}). The residual part is also scaled by a parameter $\frac{1}{k}$, where $k$ denotes the number of microstructures in LARD block. Immediately following the deep extraction unit, the upscale unit is performed to resize the LR feature maps to the size of HR images in height and width. It contains upsample layer with the $nearest$ mode and convolution layer with kernel size $3 \times 3$. Next, the global attention unit consisting of two convolution layers and two GL attention modules is added to discriminate and re-weight each feature map in the HR feature space. Each of the convolution layers contains 64 filters with the kernel size $3 \times 3$. We set stride as 1 and use zero paddings to ensure the feature maps of the same size as HR images. The parameter $s$ is set to 16 (Figure \ref{fig4}) in the GL attention module, to shrink the number of channels from 64 to 4, and to upscale it back to 64 again to obtain the global statistics for each feature map in the HR feature space. Last, the reconstruction unit with two convolution layers with 64 and 3 filters, respectively, are added in the very end of our model. These two convolution layers also have the kernel size $3 \times 3$ and are performed with stride 1 and zero paddings. The reconstruction unit eventually restores the HR outputs. 

\begin{figure}[t]
	\centering
	\begin{minipage}{0.43\textwidth}
	  \centerline{\includegraphics[width=1\linewidth]{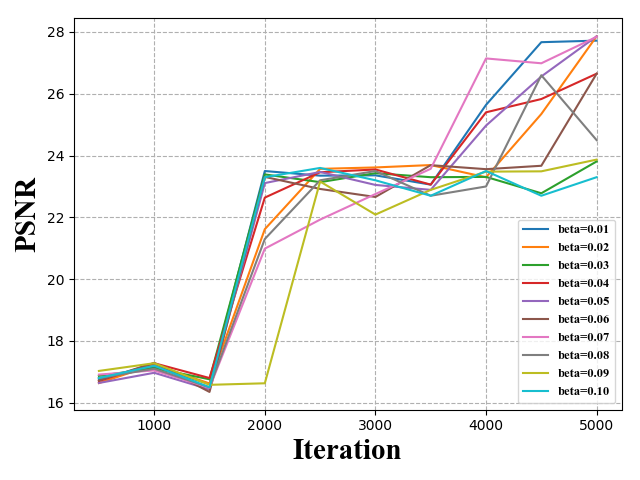}}
	\end{minipage}
	\begin{minipage}{0.44\textwidth}
	  \centerline{\includegraphics[width=1\linewidth]{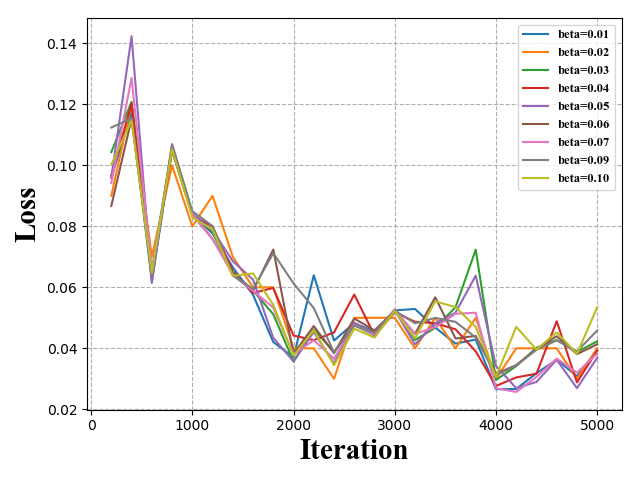}}
	\end{minipage}
	\caption{Sensitivity analysis of $\beta$} 
\label{fig6}
\end{figure}

As investigated in \cite{lim2017enhanced,nah2017deep}, the batch normalization layers get rid of range flexibility from networks by normalizing the features. So the batch normalization trends to stretch the contrast and normalize the color distribution of images, which destroys the original contrast information. Based on these analyses, we also discard the batch normalization layers in our MAANet to reduce computational complexity, increase the performance and stabilize network training. 

As shown in Eq. (10), there exists a hyper-parameter $\beta$ in our model, which controls the degree of highlighting in the LA attention module. In order to evaluate the sensitivity of $\beta$, we conduct a subset training comparison. We repeatedly train our MAANet on a subset of DIV2K dataset \cite{Timofte_2017_CVPR_Workshops}, and let $\beta$ range from 0.01 to 0.1, for each individual training. Then we compare the validation results of Set5 \cite{bevilacqua2012low} for peak signal-to-noise ratio (PSNR) and the training loss of the model. As shown in Figure \ref{fig6}, we can see that as the training goes on, the PSNR gradually increases. Moreover, the greater the iteration of training, the increasing trends of PSNR become more noticeable when $\beta$ is set to 0.07, 0.06, 0.05, and 0.04. As to the training loss, the model has better convergence ability when $\beta$ is set as 0.07, 0.02, and 0.01. Considering both the growth of PSNR and the convergence performance of training loss, we choose to set the $\beta$ at 0.07 in our model.

Our MAANet is implemented by Pytorch and trained with one NVIDIA GTX 1080Ti GPU. We use the stochastic gradient descent to optimize the loss function as shown in Eq. (6). The initial learning rate is set to $2\times 10^{-4}$, and it is evenly halved 5 times throughout the whole training process containing a total of $1\times 10^{6}$ iterations.

\subsection{Datasets and Evaluation}

\begin{table}[b]
\vspace{-0.21cm}
\renewcommand\thetable{1}
    \begin{center}
        \caption{Description of datasets}
        \label{table1} 
            \begin{tabular}{lccc}        
                \hline                   
                Dataset         & \# of Images       & Resolution          & Context        \\
                \hline
                Set5 \cite{bevilacqua2012low}            & 5             & 0.2K-0.6K          & Natural scenes              \\ 
                Set14 \cite{zeyde2010single}           & 14            & 0.2K-0.8K          & Natural scenes                 \\ 
                BSDS100 \cite{zeyde2010single}          & 100           & 0.48K              & Natural scenes                    \\
                Urban100 \cite{huang2015single}        & 100           & $\sim$1K           & Urban scenes 
                    \\
                Manga109 \cite{huang2015single}        & 109           & $\sim$1K           & Cartoon
                    \\
                \hline  
        \end{tabular}
    \end{center}
\end{table}

\begin{table*}[t]
    \renewcommand\thetable{2}
    \centering
    \begin{threeparttable}  
        \caption{Quantitative comparison with state-of-the-art competitors}  
        \label{table2}  
        \begin{tabular}{lccccccccccc}  
            \toprule  
            \multirow{2}{*}{Method}
            &\multirow{2}{*}{Scale}
            &\multicolumn{2}{c}{Set5}
            &\multicolumn{2}{c}{Set14}
            &\multicolumn{2}{c}{BSDS100}
            &\multicolumn{2}{c}{Urban100}
            &\multicolumn{2}{c}{Manga109}\cr  
            \cmidrule(lr){3-4} 
            \cmidrule(lr){5-6} 
            \cmidrule(lr){7-8} 
            \cmidrule(lr){9-10} 
            \cmidrule(lr){11-12} 
            &   &PSNR  &SSIM   &PSNR  &SSIM   &PSNR  &SSIM   &PSNR  &SSIM   &PSNR  &SSIM\cr  
            \midrule 
            Bicubic                                     &$\times 2$    &33.66 &0.9299   &30.24 &0.8688    &29.56 &0.8431   &26.88 &0.8403   &30.80 &0.9339     \cr
            SRCNN \cite{dong2016image} ($16'$)          &$\times 2$    &36.66 &0.9542   &32.45 &0.9067    &31.36 &0.8879   &29.50 &0.8946   &35.60 &0.9663     \cr
            FSRCNN \cite{dong2016accelerating} ($16'$)  &$\times 2$    &37.05 &0.9560   &32.66 &0.9090    &31.53 &0.8920   &29.88 &0.9020   &36.67 &0.9710     \cr
            VDSR \cite{kim2016accurate} ($16'$)         &$\times 2$    &37.53 &0.9590   &33.05 &0.9130    &31.90 &0.8960   &30.77 &0.9140   &37.22 &0.9750     \cr
            LapSRN \cite{lai2017deep} ($17'$)           &$\times 2$    &37.52 &0.9591   &33.08 &0.9130    &31.08 &0.8950   &30.41 &0.9101   &37.27 &0.9740     \cr
            EDSR \cite{lim2017enhanced} ($17'$)         &$\times 2$    &38.11 &0.9602   &33.92 &0.9195    &32.32 &0.9013   &32.93 &0.9351   &39.10 &0.9773     \cr
            RDN \cite{zhang2018residual} ($18'$)        &$\times 2$    &38.24 &0.9614   &34.01 &0.9212    &32.34 &0.9017   &32.89 &0.9353   &39.18 &0.9780     \cr
            DBPN \cite{haris2018deep} ($18'$)           &$\times 2$    &38.09 &0.9600   &33.85 &0.9190    &32.27 &0.9000   &32.55 &0.9324   &38.89 &0.9775     \cr
            {\bf MAANet}                         &$\times 4$    &{\bf 38.37} &{\bf 0.9618}   &{\bf 34.33} &{\bf 0.9227}    &{\bf 32.42} &{\bf 0.9027}   &{\bf 33.47} &{\bf 0.9390}   &{\bf 39.63} &{\bf 0.9788}     \cr
            \toprule 
            Bicubic                                     &$\times 4$    &28.42 &0.8104   &26.00 &0.7027    &25.96 &0.6675   &23.14 &0.6577   &24.89 &0.7866     \cr
            SRCNN \cite{dong2016image} ($16'$)          &$\times 4$    &30.48 &0.8628   &27.50 &0.7513    &26.90 &0.7101   &24.52 &0.7221   &27.58 &0.8555     \cr
            FSRCNN \cite{dong2016accelerating} ($16'$)  &$\times 4$    &30.72 &0.8660   &27.61 &0.7550    &26.98 &0.7150   &24.62 &0.7280   &27.90 &0.8610     \cr
            VDSR \cite{kim2016accurate} ($16'$)         &$\times 4$    &31.35 &0.8830   &28.02 &0.7680    &27.29 &0.7260   &25.18 &0.7540   &28.83 &0.8870     \cr
            LapSRN \cite{lai2017deep} ($17'$)           &$\times 4$    &31.54 &0.8850   &28.19 &0.7720    &27.32 &0.7270   &25.21 &0.7560   &29.09 &0.8900     \cr
            EDSR \cite{lim2017enhanced} ($17'$)         &$\times 4$    &32.46 &0.8968   &28.80 &0.7876    &27.71 &0.7420   &26.64 &0.8033   &31.02 &0.9148     \cr
            RDN \cite{zhang2018residual} ($18'$)        &$\times 4$    &32.47 &0.8990   &28.81 &0.7871    &27.72 &0.7419   &26.61 &0.8028   &31.00 &0.9151     \cr
            DBPN \cite{haris2018deep} ($18'$)           &$\times 4$    &32.47 &0.8980   &28.82 &0.7860    &27.72 &0.7400   &26.38 &0.7946   &30.91 &0.9137     \cr
            {\bf MAANet}                         &$\times 4$    &{\bf 32.85} &{\bf 0.9023}   &{\bf 29.11} &{\bf 0.7999}    &{\bf 27.83} &{\bf 0.7452}   &{\bf 26.91} &{\bf 0.8120}   &{\bf 31.73} &{\bf 0.9276}     \cr
            \bottomrule
        \end{tabular}            
        \end{threeparttable}  
\end{table*}

Our MAANet is trained on DIV2K dataset \cite{Timofte_2017_CVPR_Workshops} with 800 HR images as the training data. The DIV2K contains a broad diversity of contents at 2K resolution that provides many details. To obtain the LR counterparts for training, we downscale the HR images using bicubic and use a batch size of 16 for input LR images. As to the testing phase, we evaluate our model on five widely used benchmark datasets for SR algorithms including Set5 \cite{bevilacqua2012low}, Set14 \cite{zeyde2010single}, BSDS100 \cite{arbelaez2011contour}, Urban100 \cite{huang2015single}, and Manga109 \cite{matsui2017sketch}. These benchmark datasets cover various contents and styles, and include a wide range of resolutions. The description of these testing sets is shown in Table \ref{table1}.

As to the competitors, we compare our MAANet with 7 state-of-the-art SR algorithms including SRCNN \cite{dong2016image}, FSRCNN \cite{dong2016accelerating}, VDSR \cite{kim2016accurate}, LapSRN \cite{lai2017deep}, EDSR \cite{lim2017enhanced}, RDN \cite{zhang2018residual}, and DBPN \cite{haris2018deep}. The selection criteria for our competitors are: (1) recent work: all of these competitors are published in the most recent years; (2)DCNNs based model: all of these competitors are based on DCNNs and optimized by $L_{1}$ loss function; and (3) competitiveness: they clearly represent the state-of-the-art. All these competitors are trained on the same DIV2K dataset \cite{Timofte_2017_CVPR_Workshops} and tested on these five benchmark datasets listed in Table 1. All of them are tuned and compared based on their best performance.

\subsection{Quantitative Results}
We convert the obtained HR images to YCbCr space, and then compute the peak signal-to-noise ratio (PSNR) and structural similarity index (SSIM) on the Y channel. The PSNR is a commonly used metric to measure the similarity between two images, which is calculated using the Mean-Square-Error (MSE) of the pixels and the maximum possible pixel value. The SSIM is developed to improve traditional methods, such as PSNR, which have been proven to be inconsistent with human visual perception. It takes luminance, contrast, and structure of both images into account. The higher the PSNR and the SSIM, the better the performance of the algorithm. All the competitors and our model are evaluated on the same criteria. 

The comparison with state-of-the-art competitors is shown in Table 2. The results are for 2$\times$ and 4$\times$ super-resolution measured by PSNR and SSIM. We can see that our MAANet performs favorably against those competitors on all the evaluated datasets with all scaling factors. Moreover, it is worth noting that with the increase of scaling factors, i.e., from 2$\times$ to 4$\times$, the performance of our MAANet shows a more obvious advantage on these metrics. That is because by using the proposed LA and GA attention, our model can better deal with high-frequency information in the deep feature extraction and the global attention units. In return, it helps to restore more details from the input LR images. The results fully illustrate the superiority of our model.

\subsection{Qualitative Results}
In Figure 7, we show visual comparisons on several images extracted from testing datasets. The results are shown with a scale factor of $4\times$. It can be seen that our MAANet accurately restores the HR images with more details and higher accuracy. For example, the butterfly image is extracted from Set5. Focusing on one tiny square of size $20\times8$ pixels, we can observe that our model can better recover the actual pixel values with less difference, which endows our model a better performance on PSNR and SSIM. The YumeiroCooking image is extracted from Manga109, which is a challenging dataset containing more stripes and lines. We can see that our model perfectly recovers these stripes and lines in the HR image, and the obtained SR image by our model has much fewer blurred details compared with other competitors. The img\_076 and img\_020 images are extracted from Urban100, which is also a challenging dataset involving rich contexts of the urban environment. It can be seen that our model recovers more grid patterns and parallel straight lines.

\begin{figure*}
\centering
\subfigure{
\begin{minipage}[b]{0.23\linewidth}
\centerline{\includegraphics[width=1\linewidth]{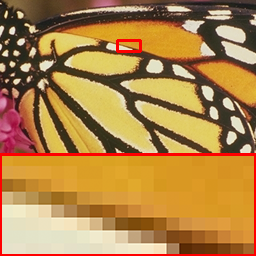}}
\centerline{HR (butterflty)}
\end{minipage}}
\subfigure{ 
\begin{minipage}[b]{0.62\linewidth}
\begin{minipage}[b]{0.24\linewidth}
\centerline{\includegraphics[width=1\linewidth]{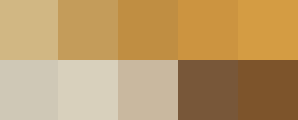}}
\centerline{LR}
\end{minipage}
\begin{minipage}[b]{0.24\linewidth}
\centerline{\includegraphics[width=1\linewidth]{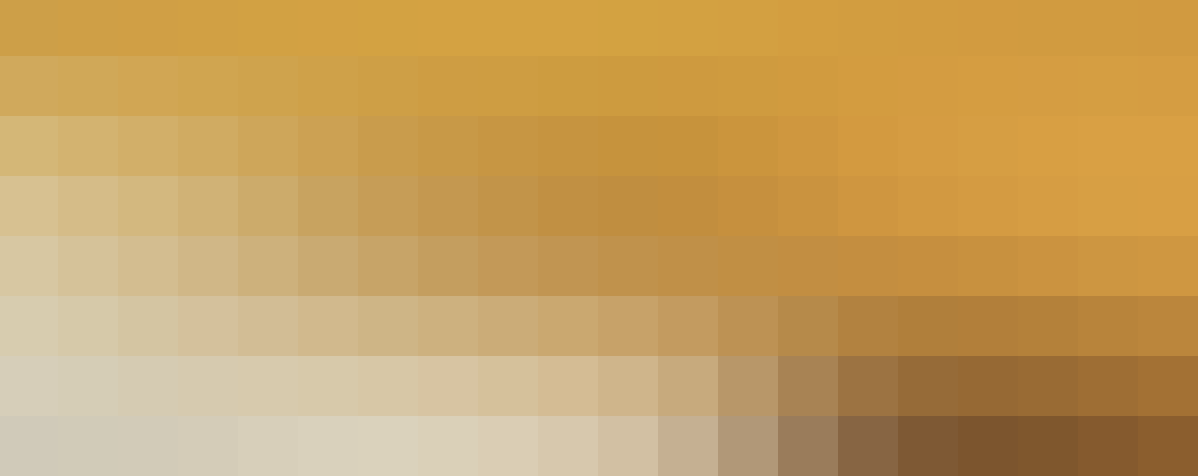}}
\centerline{Bicubic}
\end{minipage}
\begin{minipage}[b]{0.24\linewidth}
\centerline{\includegraphics[width=1\linewidth]{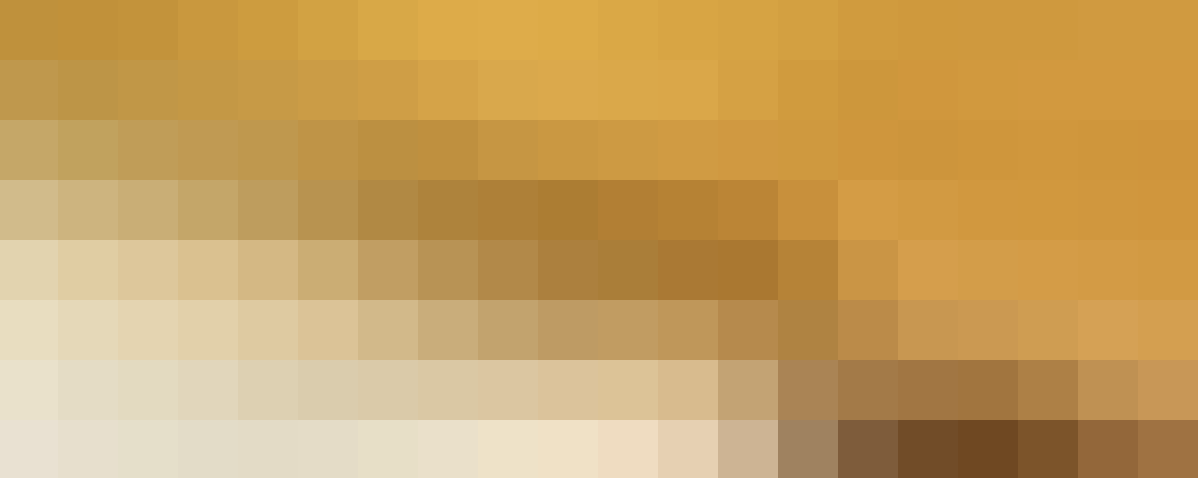}}
\centerline{SRCNN \cite{dong2016image}}
\end{minipage}
\begin{minipage}[b]{0.24\linewidth}
\centerline{\includegraphics[width=1\linewidth]{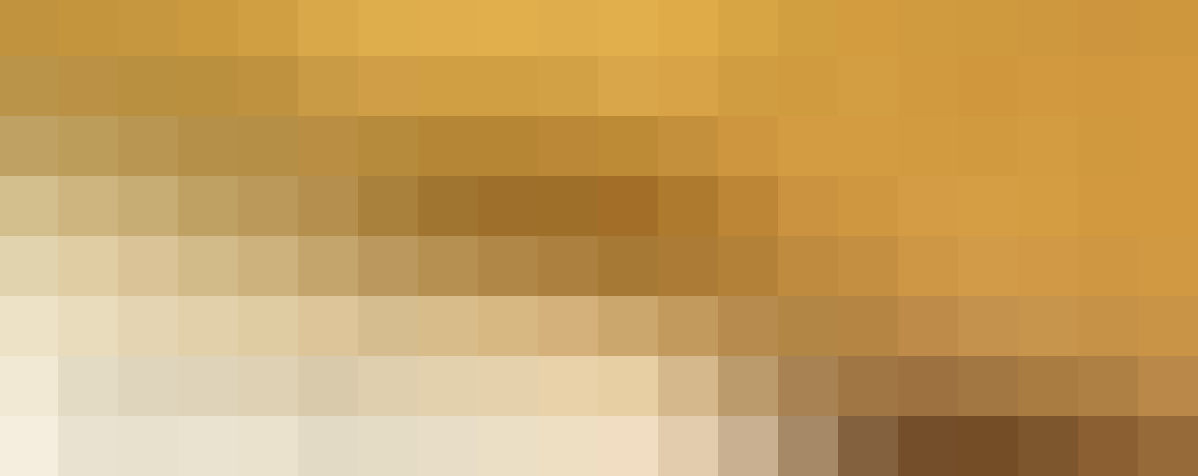}}
\centerline{FSRCNN \cite{dong2016accelerating}}
\end{minipage}\\
\begin{minipage}[b]{0.24\linewidth}
\centerline{\includegraphics[width=1\linewidth]{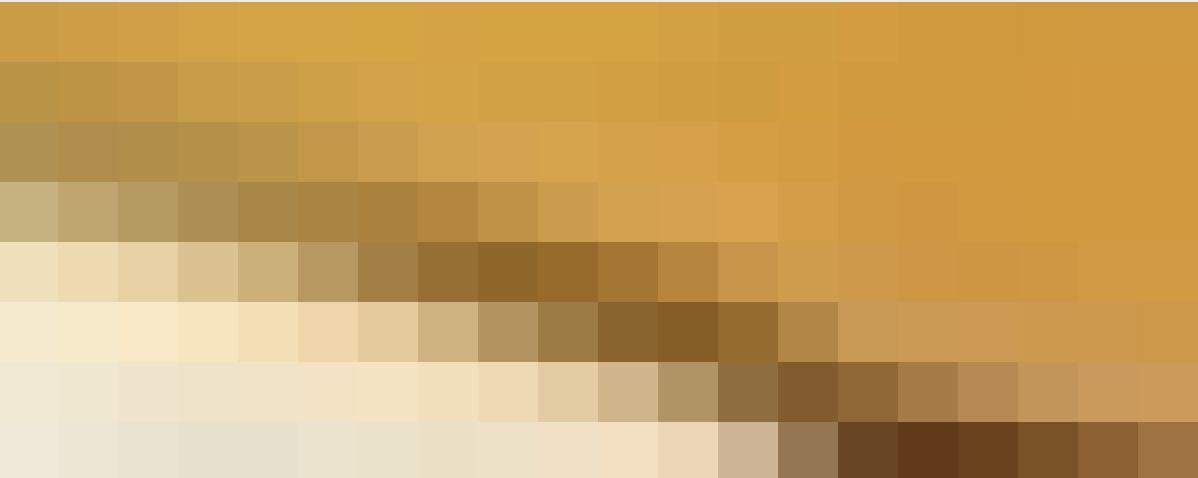}}
\centerline{VDSR \cite{kim2016accurate}}
\end{minipage}
\begin{minipage}[b]{0.24\linewidth}
\centerline{\includegraphics[width=1\linewidth]{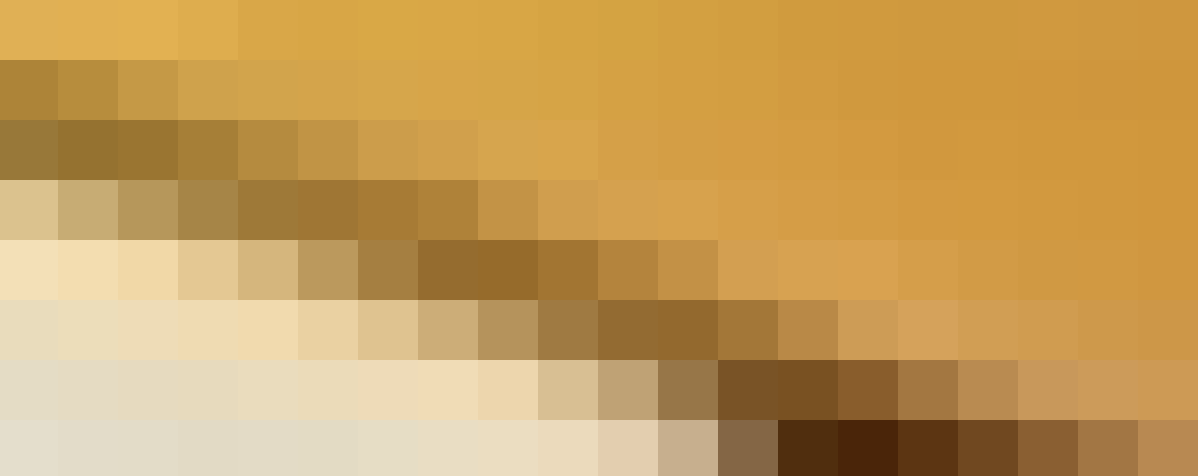}}
\centerline{LapSRN \cite{lai2017deep}}
\end{minipage}
\begin{minipage}[b]{0.24\linewidth}
\centerline{\includegraphics[width=1\linewidth]{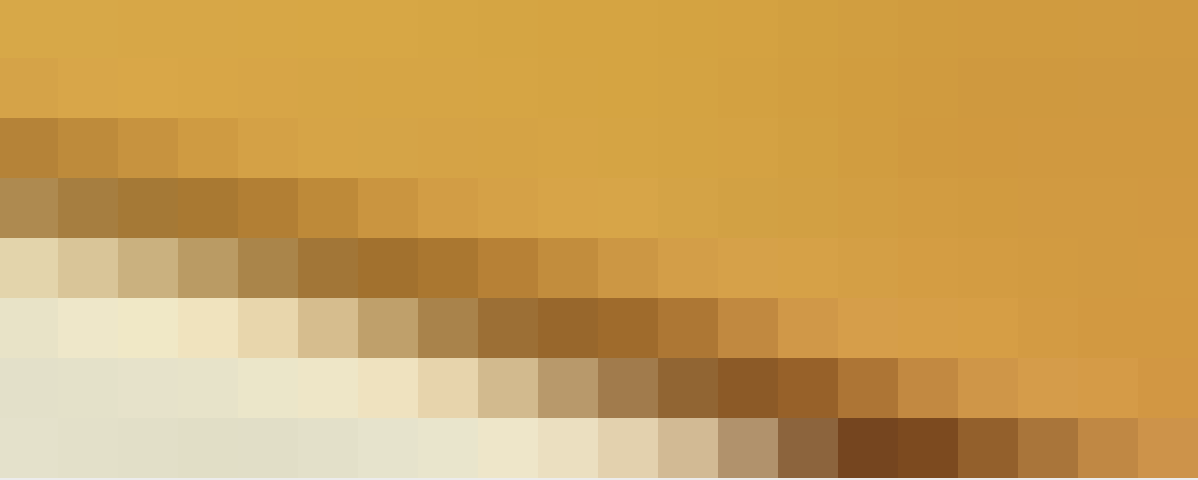}}
\centerline{EDSR \cite{lim2017enhanced}}
\end{minipage}
\begin{minipage}[b]{0.24\linewidth}
\centerline{\includegraphics[width=1\linewidth]{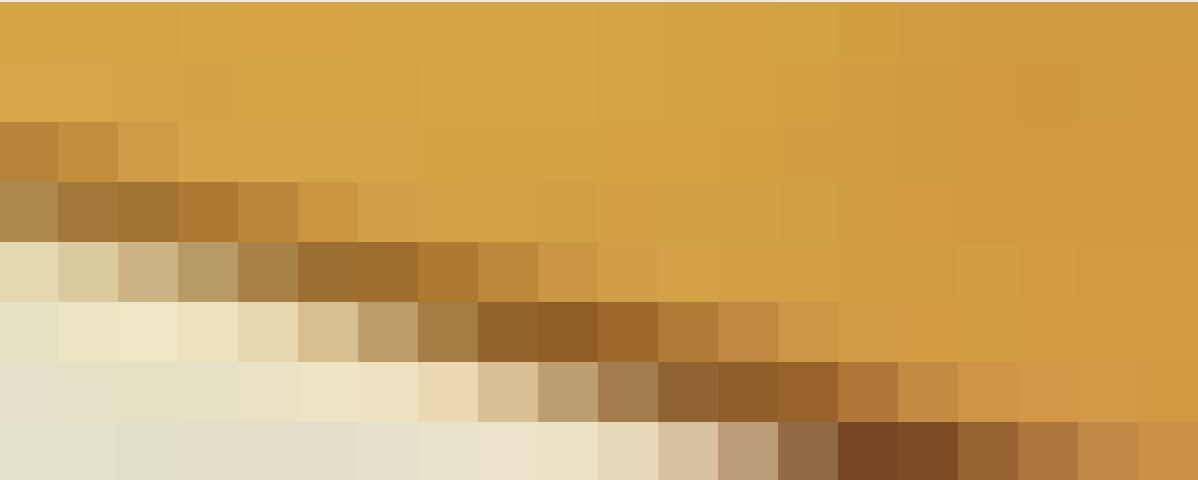}}
\centerline{RDN \cite{zhang2018residual}}
\end{minipage}\\
\begin{minipage}[b]{0.24\linewidth}
\centerline{\includegraphics[width=1\linewidth]{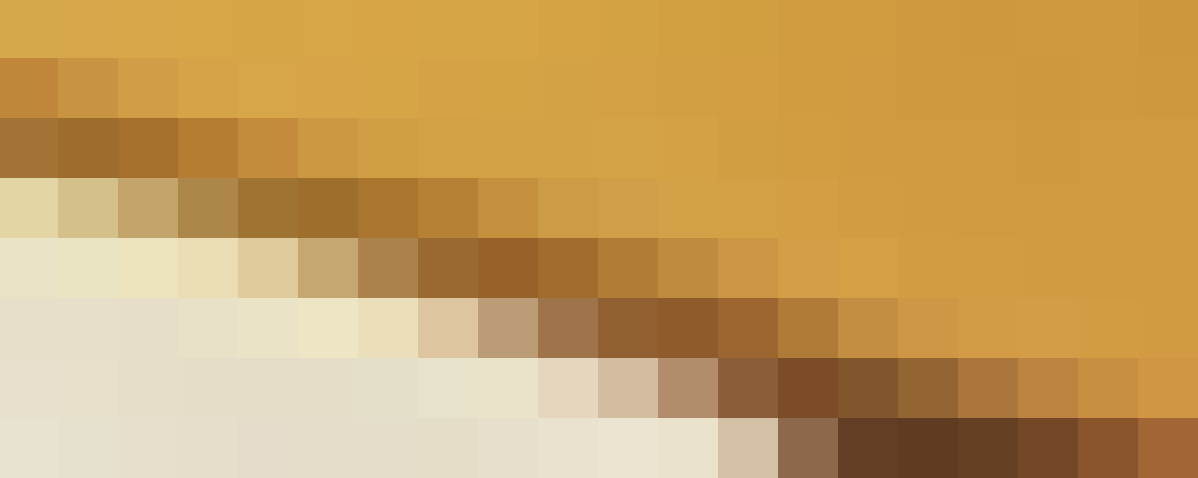}}
\centerline{DBPN \cite{haris2018deep}}
\end{minipage}
\begin{minipage}[b]{0.24\linewidth}
\centerline{\includegraphics[width=1\linewidth]{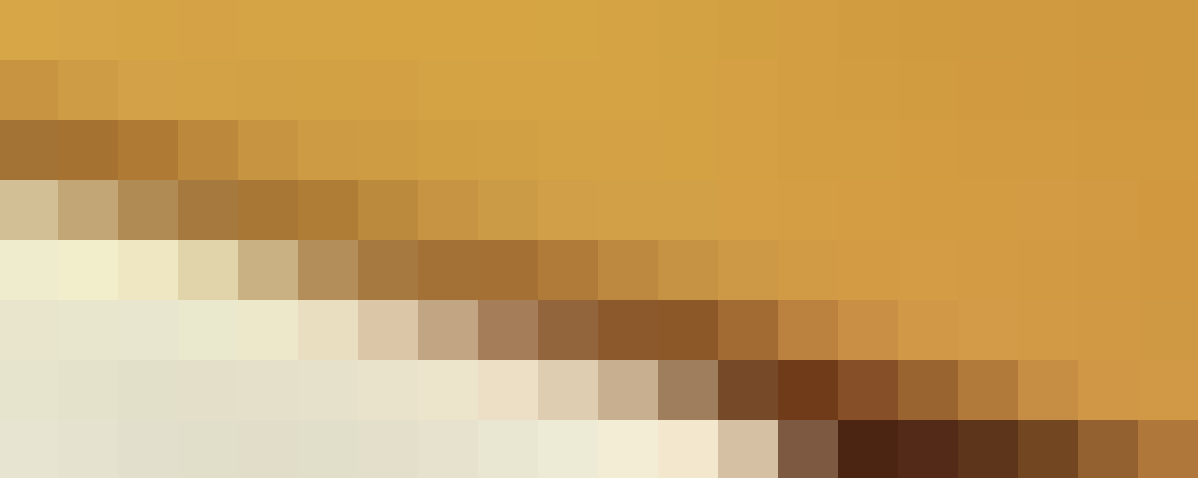}}
\centerline{MAANet}
\end{minipage}
\end{minipage}}
\subfigure{
\begin{minipage}[b]{0.21\linewidth}
\centerline{\includegraphics[width=1\linewidth]{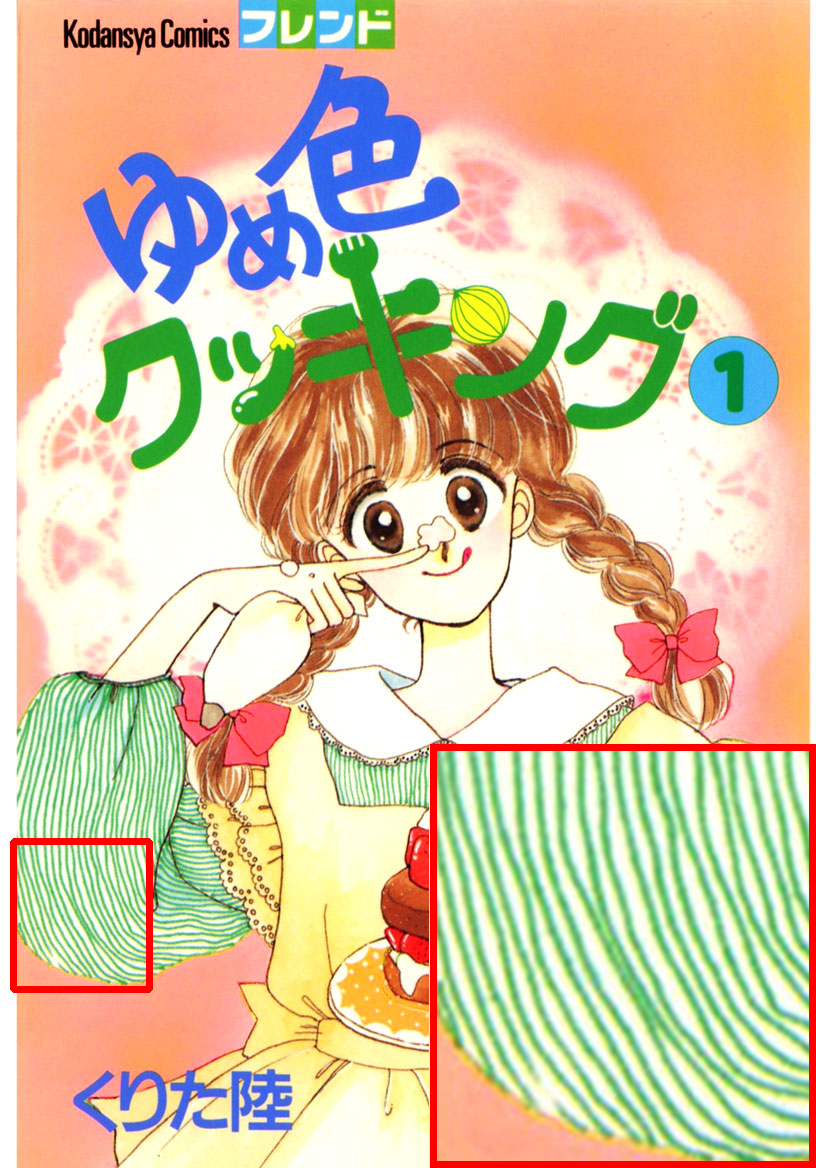}}
\centerline{HR (YumeiroCooking)}
\end{minipage}}
\subfigure{ 
\begin{minipage}[b]{0.65\linewidth}
\begin{minipage}[b]{0.19\linewidth}
\centerline{\includegraphics[width=1\linewidth]{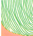}}
\centerline{LR}
\end{minipage}
\begin{minipage}[b]{0.19\linewidth}
\centerline{\includegraphics[width=1\linewidth]{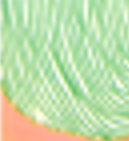}}
\centerline{Bicubic}
\end{minipage}
\begin{minipage}[b]{0.19\linewidth}
\centerline{\includegraphics[width=1\linewidth]{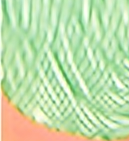}}
\centerline{SRCNN \cite{dong2016image}}
\end{minipage}
\begin{minipage}[b]{0.19\linewidth}
\centerline{\includegraphics[width=1\linewidth]{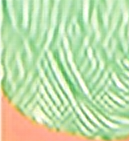}}
\centerline{FSRCNN \cite{dong2016accelerating}}
\end{minipage}
\begin{minipage}[b]{0.19\linewidth}
\centerline{\includegraphics[width=1\linewidth]{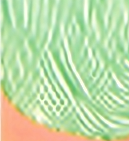}}
\centerline{VDSR \cite{kim2016accurate}}
\end{minipage}\\
\begin{minipage}[b]{0.19\linewidth}
\centerline{\includegraphics[width=1\linewidth]{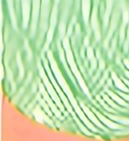}}
\centerline{LapSRN \cite{lai2017deep}}
\end{minipage}
\begin{minipage}[b]{0.19\linewidth}
\centerline{\includegraphics[width=1\linewidth]{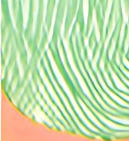}}
\centerline{EDSR \cite{lim2017enhanced}}
\end{minipage}
\begin{minipage}[b]{0.19\linewidth}
\centerline{\includegraphics[width=1\linewidth]{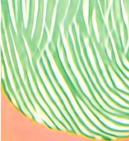}}
\centerline{RDN \cite{zhang2018residual}}
\end{minipage}
\begin{minipage}[b]{0.19\linewidth}
\centerline{\includegraphics[width=1\linewidth]{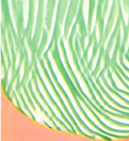}}
\centerline{DBPN \cite{haris2018deep}}
\end{minipage}
\begin{minipage}[b]{0.19\linewidth}
\centerline{\includegraphics[width=1\linewidth]{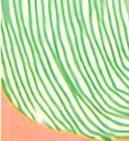}}
\centerline{MAANet}
\end{minipage}
\end{minipage}}
\subfigure{
\begin{minipage}[b]{0.27\linewidth}
\centerline{\includegraphics[width=1\linewidth]{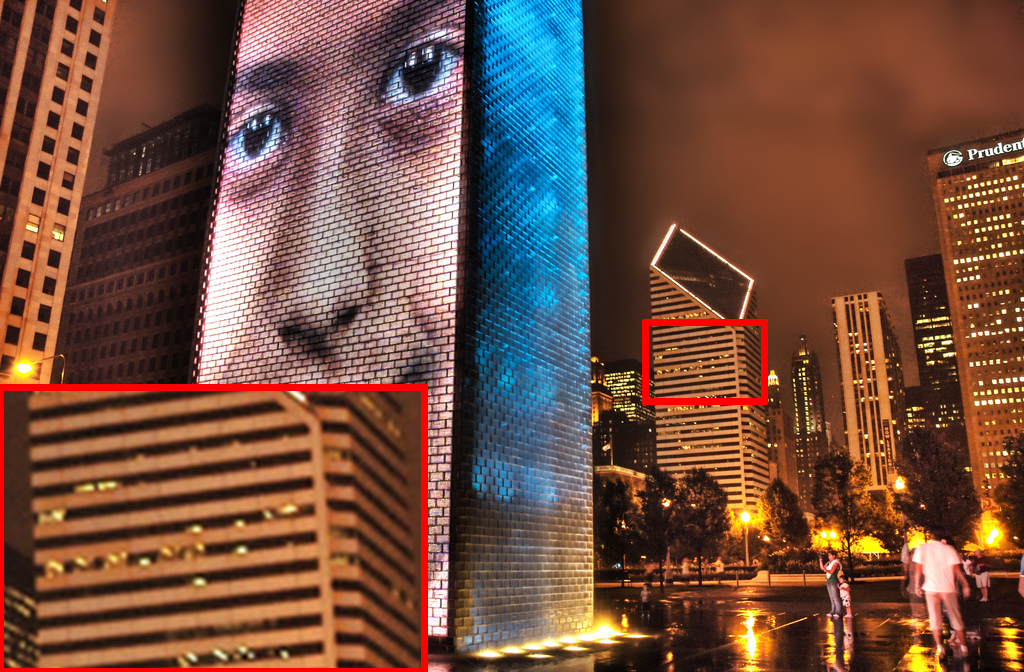}}
\centerline{HR (img\_076)}
\end{minipage}}
\subfigure{ 
\begin{minipage}[b]{0.58\linewidth}
\begin{minipage}[b]{0.19\linewidth}
\centerline{\includegraphics[width=1\linewidth]{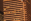}}
\centerline{LR}
\end{minipage}
\begin{minipage}[b]{0.19\linewidth}
\centerline{\includegraphics[width=1\linewidth]{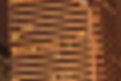}}
\centerline{Bicubic}
\end{minipage}
\begin{minipage}[b]{0.19\linewidth}
\centerline{\includegraphics[width=1\linewidth]{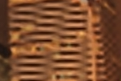}}
\centerline{SRCNN \cite{dong2016image}}
\end{minipage}
\begin{minipage}[b]{0.19\linewidth}
\centerline{\includegraphics[width=1\linewidth]{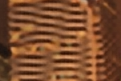}}
\centerline{FSRCNN \cite{dong2016accelerating}}
\end{minipage}
\begin{minipage}[b]{0.19\linewidth}
\centerline{\includegraphics[width=1\linewidth]{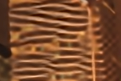}}
\centerline{VDSR \cite{kim2016accurate}}
\end{minipage}
\\
\begin{minipage}[b]{0.19\linewidth}
\centerline{\includegraphics[width=1\linewidth]{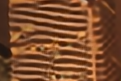}}
\centerline{LapSRN \cite{lai2017deep}}
\end{minipage}
\begin{minipage}[b]{0.19\linewidth}
\centerline{\includegraphics[width=1\linewidth]{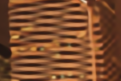}}
\centerline{EDSR \cite{lim2017enhanced}}
\end{minipage}
\begin{minipage}[b]{0.19\linewidth}
\centerline{\includegraphics[width=1\linewidth]{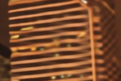}}
\centerline{RDN \cite{zhang2018residual}}
\end{minipage}
\begin{minipage}[b]{0.19\linewidth}
\centerline{\includegraphics[width=1\linewidth]{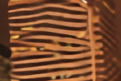}}
\centerline{DBPN \cite{haris2018deep}}
\end{minipage}
\begin{minipage}[b]{0.19\linewidth}
\centerline{\includegraphics[width=1\linewidth]{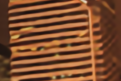}}
\centerline{MAANet}
\end{minipage}
\end{minipage}}
\subfigure{
\begin{minipage}[b]{0.35\linewidth}
\centerline{\includegraphics[width=1\linewidth]{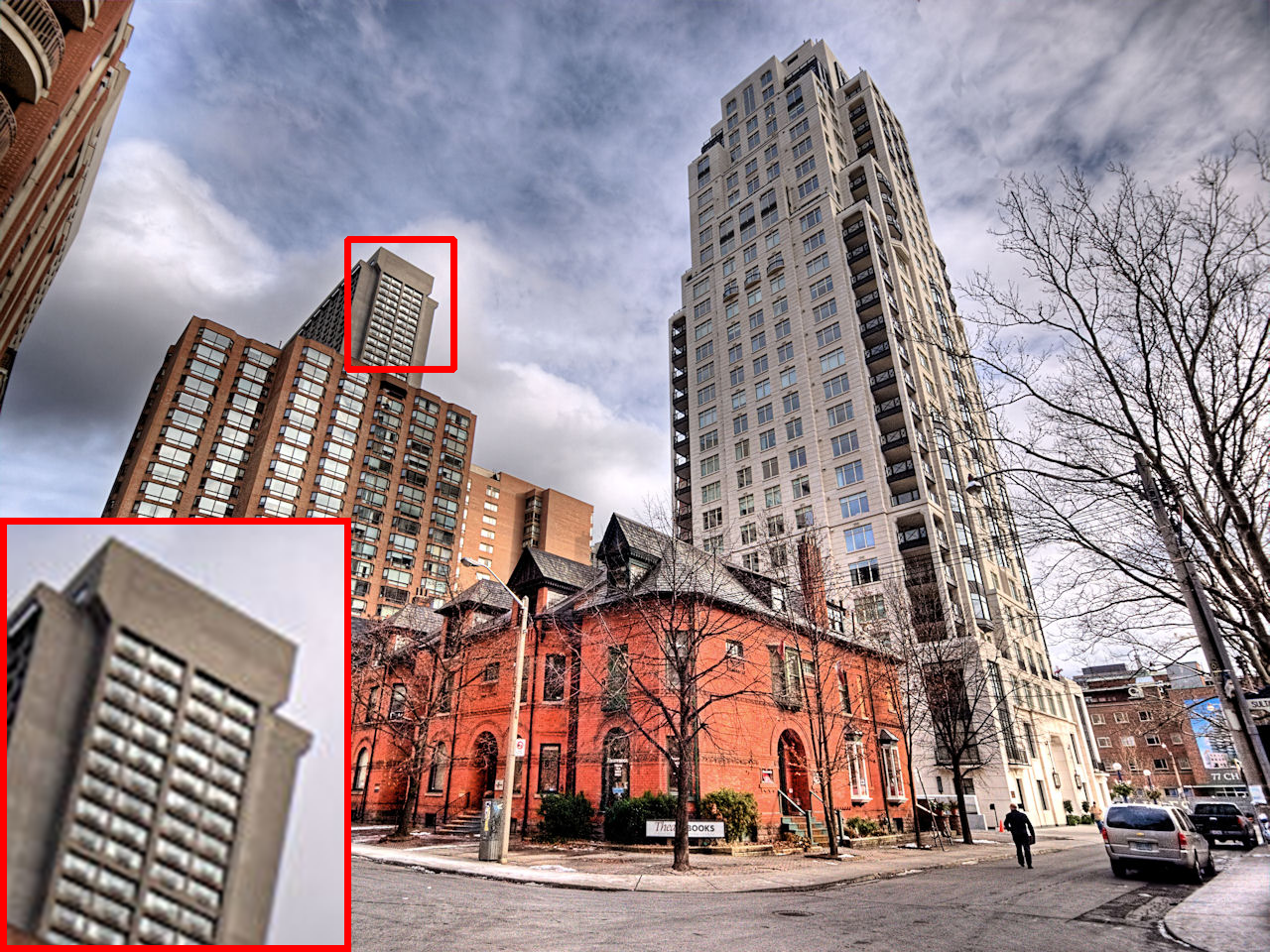}}
\centerline{HR (img\_020)}
\end{minipage}}
\subfigure{ 
\begin{minipage}[b]{0.5\linewidth}
\begin{minipage}[b]{0.19\linewidth}
\centerline{\includegraphics[width=1\linewidth]{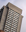}}
\centerline{LR}
\end{minipage}
\begin{minipage}[b]{0.19\linewidth}
\centerline{\includegraphics[width=1\linewidth]{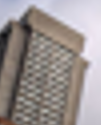}}
\centerline{Bicubic}
\end{minipage}
\begin{minipage}[b]{0.19\linewidth}
\centerline{\includegraphics[width=1\linewidth]{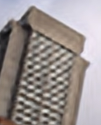}}
\centerline{SRCNN \cite{dong2016image}}
\end{minipage}
\begin{minipage}[b]{0.19\linewidth}
\centerline{\includegraphics[width=1\linewidth]{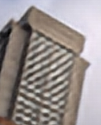}}
\centerline{FSRCNN \cite{dong2016accelerating}}
\end{minipage}
\begin{minipage}[b]{0.19\linewidth}
\centerline{\includegraphics[width=1\linewidth]{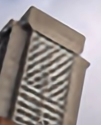}}
\centerline{VDSR \cite{kim2016accurate}}
\end{minipage}\\
\begin{minipage}[b]{0.19\linewidth}
\centerline{\includegraphics[width=1\linewidth]{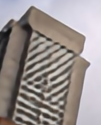}}
\centerline{LapSRN \cite{lai2017deep}}
\end{minipage}
\begin{minipage}[b]{0.19\linewidth}
\centerline{\includegraphics[width=1\linewidth]{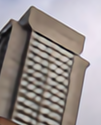}}
\centerline{EDSR \cite{lim2017enhanced}}
\end{minipage}
\begin{minipage}[b]{0.19\linewidth}
\centerline{\includegraphics[width=1\linewidth]{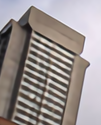}}
\centerline{RDN \cite{zhang2018residual}}
\end{minipage}
\begin{minipage}[b]{0.19\linewidth}
\centerline{\includegraphics[width=1\linewidth]{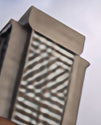}}
\centerline{DBPN \cite{haris2018deep}}
\end{minipage}
\begin{minipage}[b]{0.19\linewidth}
\centerline{\includegraphics[width=1\linewidth]{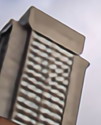}}
\centerline{MAANet}
\end{minipage}
\end{minipage}}
\caption{Visual comparisons for $4 \times$ super-resolution results. LR is the Low-resolution input and HR is the ground-truth high-resolution image.}
\end{figure*}

\section{Conclusion}
In this paper, we propose a novel image super-resolution model named Multi-view Aware Attention Networks. Our model applies the local aware and the global aware attention to deal with low-resolution images in unequal manners. Our model can highlight the high-frequency components and discriminate each feature from LR images in the local and the global views, respectively. Furthermore, we propose the local attentive residual-dense block that combines the LA attention with multiple residual and dense connections. The LARD block can be easily stacked to fit a very deep and trainable network architecture for the super-resolution task. Extensive evaluations on various benchmark datasets verified the effectiveness of our model against the state-of-the-art SR algorithms.
{\small
\bibliographystyle{ACM-Reference-Format}
\bibliography{my}
}

\end{document}